\documentclass{article} % For LaTeX2e
\usepackage{iclr2023_conference,times}
\usepackage{biblatex}
\usepackage{times}
\usepackage[utf8]{inputenc}
\usepackage{graphicx}
\usepackage{latexsym}
\usepackage{hyphenat}

\usepackage[ruled,vlined]{algorithm2e}
\usepackage{tabularx}
\usepackage{hyperref}
\usepackage{url}
\title{\nohyphens{Lexicon and Rule-based Word Lemmatization Approach for the Somali Language}}

\begin{document}

\author{Shafie Abdi Mohamed\textsuperscript{$\diamond$}, Muhidin Abdullahi Mohamed\textsuperscript{$\star$}\textsuperscript{,}\textsuperscript{$\diamond$}
\\ \textsuperscript{$\diamond$} Center for Graduate Studies, Jamhuriya University, Mogadishu, Somalia \\
\textsuperscript{$\star$} Operations and Information Management,  Aston University, Birmingham, UK \\
\hspace*{1.4ex} \textit{shafieabdi@just.edu.so}, \textit{m.mohamed10@aston.ac.uk} \\
}

\newcommand{\fix}{\marginpar{FIX}}
\newcommand{\new}{\marginpar{NEW}}

 \iclrfinalcopy

%redundency for the abstact & Introduction part-last section for example 120 experiements.
\maketitle
%https://github.com/Somali-Lemmatization/Lemmatization
\begin{abstract}
Lemmatization is a Natural Language Processing (NLP) technique used to normalize text by changing morphological derivations of words to their root forms. It is used as a core pre-processing step in many NLP tasks including text indexing, information retrieval, and machine learning for NLP, among others. This paper pioneers the development of text lemmatization for the Somali language, a low-resource language with very limited or no prior effective adoption of NLP methods and datasets. We especially develop a lexicon and rule-based lemmatizer for Somali text, which is a starting point for a full-fledged Somali lemmatization system for various NLP tasks. With consideration of the language morphological rules, we have  developed an initial lexicon of 1247 root words and 7173 derivationally related terms enriched with rules for lemmatizing words not present in the lexicon. We have tested the algorithm on 120 documents of various lengths including news articles, social media posts, and text messages. Our initial results demonstrate that the algorithm achieves an accuracy of 57\% for relatively long documents (e.g. full news articles), 60.57\% for news article extracts, and high accuracy of 95.87\% for short texts such as social media messages.% Dr halkaan maxaad ula jeedaa ma inta document oo aan ku tijaabiney
\end{abstract}

\section{Introduction}
An estimated population of over 22 million\footnote{https://www.worlddata.info/languages}, mainly in Somalia, Djibouti, Kenya, Ethiopia, the UK, the USA, and Europe speaks the Somali language. It is well known that there is no way one can understand and access information better than using one’s own language, however, there are significant digital limitations associated with the Somali language preventing it from fully participating in today’s digital world. For example, Somali is a low-resource language, which means that there are very limited or no datasets available for AI research and applications like translation, transcription, classification, language modeling, and many others. This puts the Somali-speaking people at a disadvantage and limits them from utilizing such essential AI and NLP technologies. As such, the initial research in this paper will be a starting point in meeting these limitations by pioneering the development of lemmatization for the Somali language and the construction of a lexicon dedicated to this task. 

Lemmatization is an NLP method of transforming words of the same root to their base shape. There are different methods that can be used to find the root words, for example, using rules, a dictionary-like look-up method, or a combination of the methods mentioned above. Different NLP methods and lemmatization techniques have been proposed and datasets created for many developed main languages, such as the English and French, but such resources are not available for many under-resourced languages including Somali and this is where the importance of this study comes in.

In this work, we study the problem of lemmatization in the Somali language to develop an approach for normalizing words derived from the same roots. The study mainly focuses on the lemmatization of the “MAXAA TIRI”, which is the main and written dialect of the Somali Language. The main objective of the lemmatizer is to extract meaningful root words from their inflected forms by applying look-up and rule-based methods. In other words, we built two ways to find the root words: a lexicon or dictionary based (i.e., searching words in the dictionary) and rule-based (looking at the beginning of the term and determining the root of the given word). For the lexicon-based, we constructed an initial dictionary built on the morphological association of the word lemmas and their inflected forms. Finally, the combined lexicon and rule-based methods are applied to a test dataset, achieving an average lemmatization accuracy of over 95\% for short documents. This performance shows that there is a potential for developing a full-fledged Somali Lemmatizer by scaling the lexicon and consolidating the rules. The words `dictionary' and `lexicon' are used interchangeably in the context of this paper. 

The contributions of this work are three-fold: 
\begin{enumerate}
\item First, we have developed an initial Somali lexicon for word lemmatization with the consideration of the language morphological rules.  
\item Second, we have designed a set of rules for normalizing words not covered in the dictionary and developed a Somali word lemmatization algorithm built on the lexicon and rules. 
\item Third, we have tested the algorithm on 120 documents of various lengths, including news articles and social media posts, to estimate the performance of the proposed algorithm.

\end{enumerate}

\section{Related work}

Lemmatization is a well-studied NLP task for high-resource languages including English, French, and Chinese  but remains an open research problem for under-resourced languages such as Somali~\cite{porter1980algorithm,hedderich2020survey,manjavacas2019improving}. Besides, lexicon and rule-based approaches are some of the most common methods used for word lemmatization, particularly for languages with limited resources. For example, authors in~\cite{plisson2004rule} studied the development of lemmatization algorithms based on \emph{if-then-rules} for Slovene. Similarly, Prathibha and Padma~\cite{ prathibha2015design}  put forward a rule-based lemmatization algorithm for the Kannada language by combining linguistic rules and meaningful root words. On the other hand, the work by Eger et. al.~\cite{ eger2015lexicon}  conducted a comparative study of lexicon-based approaches for Latin and found that a lexicon-based lemmatization can achieve a high accuracy given the existence of a considerable sized dataset. 

Very few studies have been conducted about the application of NLP methods in the Somali language or the construction of related datasets. In one of the very few related studies, Mohammed~\cite{ mohammed2020using} investigates the problem of part-of-speech (POS) tagging  for the Somali language using statistical and machine-learning approaches. The work reports that the introduced Somali POS tagger achieved a high accuracy of 87.51\% on tenfold cross-validation experiments. Furthermore, the work by Abdillahi, et. al.~\cite{ abdillahi2006towards} presents a study on the automatic transcription of the Somali language. Their work constructs a corpus from the audio speech of about 10 hours achieving about a 21-word error rate (WER). 

In another relevant recent study built on existing pre-trained models, Adelani, et. al.~\cite{adelani-etal-2022-thousand} investigated  the possibility of developing Machine Translation (MT) models for 16 low-resourced African languages including Somali. The study’s results suggested that a translation corpus of around two thousand sentence pairs is enough to develop MT models for low-resourced languages with a reasonable performance by fine-tuning pre-trained models. There are also several other related works that have studied approaches for mitigating the data scarcity and adoption of pre-trained models such as BERT for low-resourced languages including Somali~\cite{wang2020extending,alabi2022adapting,ogueji2021small,ogundepo2022afriteva}. 

To the best of our knowledge, there have not been any previous works on the subject of lemmatization for the Somali language, which makes this work the first of its kind for normalizing text written in Somali. It, therefore, paves the way for further research on the creation of NLP resources and techniques for the Somali language.  

\section{Proposed lemmatization approach}
\label{sec:length}
\begin{figure}[ht!]
\centering
\includegraphics[width=0.78\textwidth]{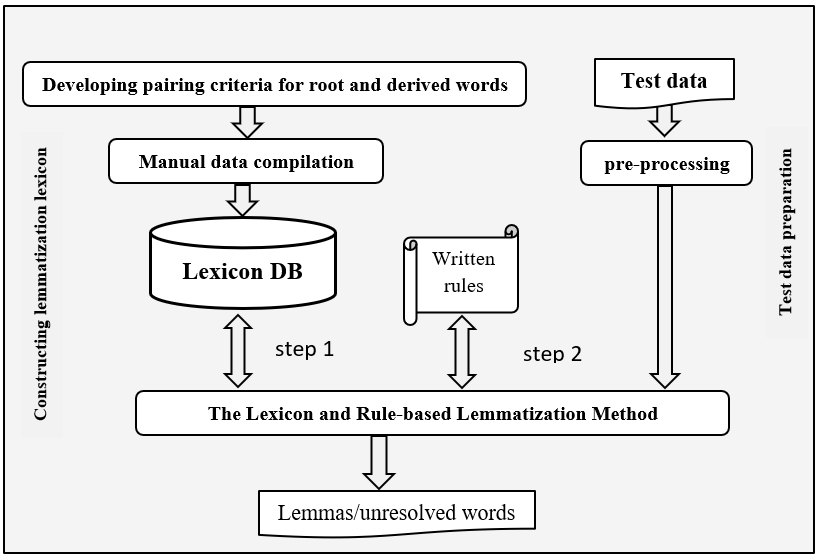}
\caption{Proposed lexicon and rule-based Somali word lemmatizer}
\label{TheLemmatizer}
\end{figure}

Figure~\ref{TheLemmatizer}
provides a high-level illustration of the proposed Somali language word lemmatizer. As shown, constructing the lexicon is the first step of the method in which criteria for pairing root and derived words were developed before commencing the actual compilation of the lexicon. The lexicon is organized into key-value pairs with the keys representing root words and values constituting the derivational forms of the roots. Because of the limited size of the lexicon created in this initial work, we have also developed simple rules for handling words that are not found in the lexicon to improve the lemmatizer accuracy. In other words, the proposed lemmmatization is a two-stage process (denoted by step 1 and step 2 in Figure~\ref{TheLemmatizer}). In the first stage, the algorithm attempts to lemmatize a given word by searching for it in the constructed lexicon and applies the written rules in stage 2 if the root word is not found in the dictionary.

%The numbers between the lexicon/written rules and the lemmatization method in the figure indicate the lemmatizer's order to normalize text. In other words, it first attempts to find the lemma from the lexicon and applies the written rules if the root word cannot be found in the dictionary. 

The lemmatization proposed in this work is particularly for the main Somali dialect, “MAXAA TIRI”. To perform the lemmatization, we first read the prepared lexicon database. The loaded data is  then converted to a JSON format to enable us to further process the lexicon and retrieve the root words. Before extracting the lemmas, we applied some data pre-processing such as removing punctuation marks and stop-words as the target of the lemmatizer is the keywords.  Finally, we look for root words in the JSON-formatted lexicon. If the base form of a given word is found in the lexicon, it is returned as the root. For words without entries in the dictionary, the rule-based component of the algorithm is applied to lemmatize the word. If the word is still not normalized at this stage, it is recognized as \emph{unsolved}, i.e, the algorithm could not lemmatize it. 
%"dowlad":["dowladda","dowlada","dowlad","dowladdeena"
\subsection{The Lexicon}
The authors, who are native Somali speakers, compiled an initial collection of the most frequently used words in several domains such as news, sport, and social media. Next, we conducted consultations with language experts at the Somali Language Academy\footnote{https://aga.so/so/}, a regulating body for the development and promotion of the Somali language. From these consultations with the experts at the academy, a systematic method for constructing pairs of root and derivative words was put in place, taking into account the language’s grammar and morphology. The collected corpus of words was predominantly of two parts: verbs and  nouns. Based on the Somali language morphology, most verbs can be changed to their root words by returning them to the command form of the verb, for example, the terms \emph{``cabay"} (drunk, past), \emph{``cabaya"} (drinking, continuous), \emph{``cabid"} (drink, noun) can all be converted to the root word \emph{``cab"} (drink now, command). Other similar roots such as  \emph{``cun"} (eat now), \emph{``laq"} (swallow now), can be obtained in the same way. With the noun category, the origin of the noun is the singular form,  for example, the words \emph{``dowladda"} (the government), \emph{``dowladdeena"} (our government) can be reduced to the lemma \emph{``dowlad"} (government). Each word may have different forms with distinct meanings different from the root word, for example, let's see the lexicon entries for the root words, \emph{``cab"} and \emph{``bariis"} with their derivatives:  

\emph{\textbf{Example 1}: entry for the root verb ``cab" (drink): cab:[``caba",``cabay",``cabeen",``caba",``cab"]}

\emph{\textbf{Example 2}: entry for the root noun ``bariis" (rice): bariis:[``bariis","bariiska"]}  

\begin{figure}[ht!]
\centering
\includegraphics[width=0.75\textwidth]{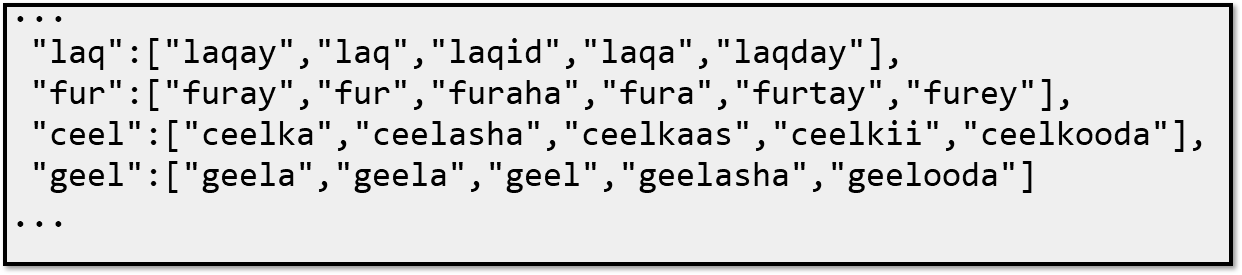}
\caption{A sample extract from the Somali lemmatizatiom lexicon}
\label{Lexcicon_extract}
\end{figure}

You may notice that word derivations are reflected with changes in the suffixes. We have built the lexicon in the form of a key and values as illustrated in the above examples (cf. the extract in Figure~\ref{Lexcicon_extract}). This structure helps in the search for the root words, i.e., if the key is found in the entry, it is returned as the root reducing the time complexity of the algorithm. Constructed in those pairs of root and derived words, the lexicon dataset is stored in a machine-readable dictionary, which is easily searchable and expandable if new terms and their derivations are to be added. At the time of writing this paper, the final lexicon consisted of a total of 8420 tokens including 1247 root words and 7173 derivatives. As an illustration of the overall lexicon organization, Figure~\ref{Lexcicon_extract} shows an extract of our lemmatization lexicon including the words: \emph{``laq"} (swallow), \emph{``fur"} (open), \emph{``ceel"}(well), \emph{"geel"} (camel), and their morphological derivations.

%Tusaale Table ah iyo Formula la raacay
\subsection{The Rules}
To expand the lemmatized vocabularies, we have adopted a rule-based method in which the root is determined from word beginnings. If different inflected forms of the same base word start with a specified sequence followed by several possible suffixes,  then a rule can be built around this to return the root word. For example, the regular expression \textbf{``$jil(c|ic|eec)\backslash w*$"}\footnote{match the sequence \emph{`jil'} followed by \emph{`c'}, \emph{`ic'} or \emph{`eec'}, then by other letters} (or equivalent \emph{``if-then-else"} rule) is used to lemmatize the inflected forms of the word ``jilci" (which means `clarify' or `soften' in Somali). Algorithm 1 summarizes the base word identification process as proposed in this work. 

\begin{algorithm}[hbt]
 \caption{Get root words from the supplied lexicon or rules}
 \label{Algo:2}
% \label{Algo:1}
% \algsetup{linenosize=\medium}
\scriptsize
\SetAlgoLined
\SetKwFunction{}
    ((i)  Input: Load each sentence;\\
    (iii) Perform word tokenization\\
    (iv) Remove stop \& duplicate words, and punctuation marks;\\
    (v) \For{each word}{
         \For{key and value in lexicon}{
           \tcc{search a specific word in the lexicon}
                \If{word is found in the lexicon}{
                  return the key in that lexicon entry as the root\\
                  update the corresponding statistic\\
                } 
               \tcc{find the roots for words not found in the lexicon using rules}
                \If{word match is found in the list of rules}{
                        return the specified root \\
                        update the corresponding statistic;
                 }
                 
                 \Else{Error: unresolved word}
               }}
\vspace{-4mm}
\end{algorithm}
The pseudocode  in Algorithm 1 is complementary to Figure~\ref{TheLemmatizer} in summarizing how the root word look-up and application of the lemmatization rules work in this proposal. Input texts are first pre-processed (e.g. tokenizing, removing punctuation marks and stop words, etc) before searching the words in the lexicon or with the rules. Words that can neither be found in the lexicon nor normalized with the designed rules are classified as \emph{unresolved} words.

\section{Experimental Evaluation}% Methodolgy
%As mentioned previously, this study is done for Somali texts written in the “MAXAA TIRI” (aka Somali-Somali). 
As previously mentioned, the proposed lemmatization method is built on the created lexicon of 1247  root words and  7173 related inflectional forms, and a set of enriching rules. The number of derivational words associated with each root in the lexicon ranged from 2 to more than 20 derivative words. Some of the primary challenges faced in relation to the lexicon construction include the lack of any existing morphological data for the Somali language and the difficulty of objectively judging the quality of the lemmas and their derivatives. 

Empirically speaking, to find the root for a given word in a text, a simple lexicon search is applied in the first instance. If the word is not found in the dictionary, the next step is to search for it in the list of designed lemmatization rules. A user of this algorithm will normally supply a text to be passed to the algorithm after which the text will be pre-processed and normalized with the lemmatizer. The primary metric utilized to assess the effectiveness of the suggested lemmatizer is accuracy. The measure is calculated as the ratio of successfully lemmatized words to all the different input-inflected words. The following equation shows the calculation of the lemmatization accuracy in this work.

\begin{equation}
Accurracy(\%) = \frac{Total\;words\;resolved}{document\;length}
\end{equation}

%\subsection{Results and discussion}
Next, we present sets of tests to evaluate the effectiveness of the proposed lemmatization method in terms of the aforementioned accuracy measure.  In this initial evaluation, we collected 120 test documents of different sizes, largely from the news and social media domains. The test dataset is then pre-processed by tokenizing it, and removing punctuation marks, stop words, and duplicates.
%In other words, the lexicon we have created has reached up to 1127 words in terms of roots, but each word has many derivational words. Sometimes, it is possible for each word to generate more than 20  derivative words. From the initially constructed data, the derivative words reached 13524.
Let's use  an example sentence to illustrate how the proposed lemmatizer works (numbering of the example continues from the previous examples used in the paper). 

\emph{\textbf{Example 3 (Somali)}: Waxaan kula taliyey inuu casriyeeyo xirfadihiisa shaqo.}

\emph{\textbf{Translation (English)}: I have advised him to update his job skills.}

\begin{figure}[ht!]
\centering
\includegraphics[width=0.85\textwidth]{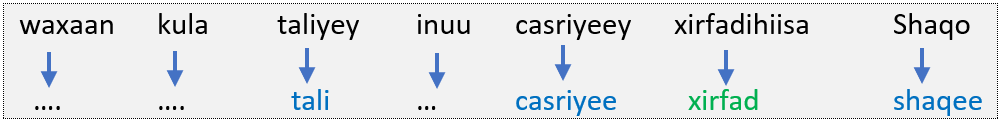}
\caption{Lemmatizing the words in the illustrative example (Example 3)}
\label{Illustrate_example}
\end{figure}

% \begin{table}[htbp]
% {\begin{tabular}{|m{13.5cm}|}
% \hline
% Waa Tartanka Hurdada oo Lagu Qabta Koonfurta Kuuriya Hal Saac Hadaad Hurudo Waxaa Lagu
% Siinayaa 1,400\$ Qofba Qofka uu Ka Hurdo Badanyahayna Waa uu kasii Lacag Badanyahay.\\ Waa Mid Kamid ah Tartamada ugu Yaabka Badan aduunka Adiga\\ Imisa/Meeqo saac baad Hurdi Kartaa? Abdifatah Haji.\\
% \hline
% \end{tabular}}
% \captionof{Figure 2 }{:Short Illustrative document}\label{ExQuery}
% \end{table}

% The document in Figure 2 is one of the tests we have done, although the length of this one is relatively small. There are other documents that are larger than this in the test set.

Figure~\ref{Illustrate_example} demonstrates the process of normalizing the words in Example 3. The first step of applying the lemmatizer to test data, such as this example, is to pre-process the text (see Figure~\ref{TheLemmatizer}). As part of this pre-processing, the terms \emph{``waxan"}, \emph{``kula"}, \emph{``inuu"} are recognised as stop words and removed. Similarly, the dot (``.") is identified as a special character and eliminated. The blue coded words (\emph{``tali", ``casriyee", ``xirfad"}) are the root words obtained from the lexicon for the corresponding terms while the word \emph{``xirfad"} (green) was not found in the lexicon but resolved with the rule-based component of the algorithm. 
\begin{table}[hbt!]
\caption{\label{tab:summarystats} Summary statistics of Example 3 lemmatization}
{\begin{tabular}{m{8cm} m{5cm}}
\hline
    \textbf{Statistic} & \textbf{Values}\\
     \hline
     Original document size in words & 8\\
     %\hline
     Stop words(non-unique) & 3\\
     %\hline
      Special characters & 1\\
     %\hline
       Unresolved words & 0\\
     %\hline
       Resolved words & 4\\
     %\hline
        Percent found & 100\% \\
     %\hline
        words from lexicon & 3\\
     %\hline
        words from rule-based & 1\\
     \hline
\end{tabular}}
\end{table}

The lemmatization summary statistics of the Example 3 sentence are also provided in Table~\ref{tab:summarystats}. In this case, the percentage of words that were normalized for the example reached 100\%, which means that all content words (excluding stop words and special characters) are lemmatized. This may be due to the fact that this is a short document, a sentence of 8 words. Unlike the lemmatization statistics of this example, a proportion of words in any typical text document (i.e., longer than a sentence) will normally remain unresolved - words that the algorithm fails to lemmatize in both stages.
 
 Overall and as part of evaluating the proposed method, we have tested the algorithm on 120 documents of various lengths including general news articles, and social media posts. For the news articles, we have used extracts (i.e., title and first 1-2 paragraphs) as well as the full articles to see the effect of document length. The results we found for these different document categories are summarized in Table~\ref{test_results}. The notations $\# Docs$, $Avg\:Doc\:Len$, and $Avg\:Acc.$ in the table respectively represent the number of documents, average document length in words, and average lemmatization accuracy. As shown, the results demonstrate that the algorithm achieves a relatively good accuracy of 57\% for moderately long documents (e.g. news articles), 60.57\% for news article extracts, and high accuracy of 95.87\% for short texts such as social media messages. This initial result is promising for a language with extremely limited or no mentionable previous NLP resources.
\begin{table}[h]
%\resizebox{\columnwidth}{!}
\caption{Test results  of the lemmatization algorithm} 
{\begin{tabular}{ m{4cm} m{2cm} m{3cm} m{3cm} m{3cm} }
    \hline
    Type &  \# Docs & Avg Doc Len & Avg Acc.\\
     \hline
     News (extract) & 75 & 74.63 & 60.57\%\\
      %\hline
      News (full) & 10 & 175.9 & 57\%\\
     %\hline
      Social media & 30 & 21.23 & 95.87\%\\
      %\hline
      Text messages & 5 & 25.2 & 68.59\%\\
         \hline
\end{tabular}}
\label{test_results}
\end{table}
\\From these results, it can be thought that the developed lexicon and the rule-based lemmatization approach works best  for short text documents such as the social media messages. From the results (cf. Table~\ref{tab:summarystats}), we can also notice that most words are normalized at the dictionary look-up level, particularly when it comes to short and moderate-sized texts. Finally, the implementation details including the source code,  lexicon, and  the collection of used test datasets are made publicly available for the research community on this GitHub page~\footnote{\url{https://github.com/ShafieAbdi/SomaliLemmatizer}}.
% \begin{table}
% \centering
% \caption{\label{tab:table-name} above table abbreviated word description}
% {\begin{tabular}{ m{4cm} m{4cm}}

%     \hline
%     abbreviated word &  \ full form\\
%      \hline
%      # docs & number of documents\\
%       %\hline
%       Avg Doc Len & average document length\\
%      %\hline
%       Avg Acc & average accuracy\\
%          \hline
% \end{tabular}}
% \end{table}\
\section{Conclusion and future work}
This work is the first of its kind to study  text lemmatization for the Somali language. It proposes a lexicon and rule-based lemmatizer for Somali written texts, which can be considered a starting point for a full-fledged Somali lemmatization system and an indispensable component of other NLP tasks. With the consideration of the Somali language morphology, we have manually crafted an initial lexicon of 1247 roots and 7173 derived words to serve as a dictionary look-up database for the word lemmatization task. Furthermore, we have built two ways to find the root words: a lexicon-based look-up method (achieved by searching in the dictionary), and a rule-based (identifying roots based on matches of word beginnings). Finally, we have performed initial evaluation of the algorithm and found some promising results including an average lemmatization accuracy of over 95\% for short documents. This means that the study has reached a reasonable level of accuracy, paving the way for further research  in enabling the Somali language with NLP techniques.

Since this research is still in progress and represents a starting point for an important NLP task and the development of underpinning resources, it has its own limitations. Therefore, our immediate future plan is to further improve the proposed lemmatization approach by increasing the size of the lexicon database and expanding the applied rules. We will also investigate the possibility of devising a methodology for building the lemmatization lexicon automatically from a collected Somali corpus. 

% First, we thank the almighty Allah for enabling us to do research in this exciting field. Secondly, We would like to express our gratitude and thankfulness to the Somali Academy Of Science And Arts for spending a lot of time with us and helping us check the accuracy of the Somali language words. Finally, we thank everyone who volunteered to contribute to this work.
%\newpage
\printbibliography
%\bibliography{latex/iclr2023_conference}

\end{document}